\documentclass[10pt,twocolumn,letterpaper]{article}

\usepackage{cvpr}              % To produce the CAMERA-READY version
\usepackage{graphicx}
\usepackage{amsmath}
\usepackage{amssymb}
\usepackage{booktabs, multirow}
\graphicspath{ {figures/} }

\usepackage{threeparttable}
\usepackage[pagebackref,breaklinks,colorlinks]{hyperref}

\usepackage{color,soul}

\newcommand{\bbR}    {\mathbb{R}}

\usepackage[capitalize]{cleveref}
\crefname{section}{Sec.}{Secs.}
\Crefname{section}{Section}{Sections}
\Crefname{table}{Table}{Tables}
\crefname{table}{Tab.}{Tabs.}

\def\cvprPaperID{*****} % *** Enter the CVPR Paper ID here
\def\confName{CVPR}
\def\confYear{2023}

\begin{document}

\title{Transformer Encoder with Multiscale Deep Learning for Pain Classification Using Physiological Signals}

\author{Zhenyuan Lu\\
{\tt\small lu.zhenyua@northeastern.edu}\\
Northeastern University\\
Boston, MA\\
\and
Burcu Ozek\\
{\tt\small ozek.b@northeastern.edu}\\
Northeastern University\\
Boston, MA\\
\and
Sagar Kamarthi\\
{\tt\small s.kamarthi@northeastern.edu}\\
Northeastern University\\
Boston, MA
}
\maketitle

%%%%%%%%% ABSTRACT
\begin{abstract}
  Pain is a serious worldwide health problem that affects a vast proportion of the population. For efficient pain management and treatment, accurate classification and evaluation of pain severity are necessary. However, this can be challenging as pain is a subjective sensation-driven experience. Traditional techniques for measuring pain intensity, \eg self-report scales, are susceptible to bias and unreliable in some instances. Consequently, there is a need for more objective and automatic pain intensity assessment strategies. In this paper, we develop PainAttnNet (PAN), a novel transformer-encoder deep-learning framework for classifying pain intensities with physiological signals as input. The proposed approach is comprised of three feature extraction architectures: multiscale convolutional networks (MSCN), a squeeze-and-excitation residual network (SEResNet), and a transformer encoder block. On the basis of pain stimuli, MSCN extracts short- and long-window information as well as sequential features. SEResNet highlights relevant extracted features by mapping the interdependencies among features. The third module employs a transformer encoder consisting of three temporal convolutional networks (TCN) with three multi-head attention (MHA) layers to extract temporal dependencies from the features. Using the publicly available BioVid pain dataset, we test the proposed PainAttnNet model and demonstrate that our outcomes outperform state-of-the-art models. These results confirm that our approach can be utilized for automated classification of pain intensity using physiological signals to improve pain management and treatment. The source code and other supplemental information are documented: \url{https://github.com/zhenyuanlu/PainAttnNet}.
\end{abstract}

%%%%%%%%% BODY TEXT
\section{Introduction}
\label{sec:intro}
Pain is a distressing sensation and emotional experience that is associated with potential or actual tissue damage in the body \cite{merskey1979pain}. It serves the purpose of alerting the body's defense mechanism to react to a stimulus to prevent further harm. Pain can seriously affect one's physical, mental, and social well-being \cite{breivik2006survey, tsang2008common}. According to the World Health Organization (WHO), those with chronic pain are more than twice as likely to have problems functioning and four times more probable to suffer from depression or anxiety \cite{gureje1998persistent}. In addition, the International Association for the Study of Pain (IASP) also estimates that 20\% of adults worldwide experience pain daily and that 10\% of adults are formally diagnosed with chronic pain each year \cite{international2012unrelieved}. 

The most prevalent criteria for categorizing pain are based on: (1) the pathophysiological mechanism (nociceptive or neuropathic pain from tissue or nerve injury, respectively), (2) the pattern of duration (\eg, acute, chronic, recurring), (3) the anatomical location involved (\eg, neck, back, knee), or (4) the etiology (\eg, malignant associated with cancer, pinched nerve, dislocated joint) \cite{woolf1998towards, thienhaus2002classification, yam2018general, abdelsayed2019different}. Therefore, pain is an essential indicator, which can range in intensity from mild to severe, that something is wrong with the body and thus can drive a person to seek medical care. Over the past two decades, pain research has been an increasingly popular field of study. The authors of this paper investigated and analyzed 264,560 research articles published since 2002 on the topic of pain using a keyword co-occurrence network (KCN) architecture \cite{ozek2022review}. According to this study \cite{ozek2022review}, there has been a sevenfold increase in the use of “pain” as a keyword and a near doubling in the number of papers discussing pain in the scientific literature.

To enhance one's health and quality of life, it is essential to gain insight into pain and develop effective pain management strategies \cite{katz2002impact, AzizabadiFarahani2010}. One of the significant obstacles to effectively manage pain is the lack of appropriate pain assessment \cite{anderson2000minority}. Proper pain assessment is also necessary for both tracking the effectiveness of pain management strategies and monitoring changes in pain intensity over time. Pain assessment techniques help clinicians and researchers to identify the causes of pain, develop new treatments, and improve our understanding of how the body processes and responds to pain \cite{wells2008improving}. Therefore, accurate pain assessments are vital for effective pain management, as it enables healthcare providers to determine the most suitable treatments for each individual  \cite{leigheb2017prospective}.

The most well-known method to assess pain intensity for individuals is using self-report scales, \eg the verbal rating scales (VR), Visual Analog Scale (VAS), or the Numeric Rating Scale (NRS), which rely on subjective self-assessment \cite{LAZARIDOU201839}. Despite these scales can offer valuable information on a person's pain experience, the pain assessment can be challenging in certain populations, \eg neonatal infants \cite{blount2009behavioural, ERIKSSON2019101003, cascella2019challenge} and individuals with cognitive impairments or communication difficulties \cite{deldar2018challenges, Werner2022}. As a result, there is a need for more objective and automated methods for assessing pain intensity \cite{Ghada2018}.

One common approach to meeting this need is the use of physiological signals, \eg electrodermal activity (EDA), electrocardiography (ECG), electromyography (EMG), and electroencephalography (EEG), to classify pain intensity \cite{Werner2022}. EDA (also known as the galvanic skin response (GSR)) detects variations in the skin conductance level (SCL), which closely corresponds with sweat gland activation. In clinical settings, skin conductance has also been employed as a substitute for pain \cite{ledowski2009monitoring}. The EDA complex comprises of sympathetic neuronal activity-generated tonic (known as skin conductance level, SCL) and phasic (known as skin conductance response SCR) components \cite{braithwaite2013guide}. ECG captures the electrical activity of the heart in order to assess cardiac health and stress. EMG monitors muscle activity and identifies changes in muscular tension, whereas EEG examines the brain's electrical activity. These signals can be used to investigate the effectiveness of pain management and shed light on how the body reacts to pain. It is possible to use them in combination with other tools to get a broader picture of the level of pain being experienced \cite{Erekat2021}. In recent years, EDA signals for pain intensity classification have gained popularity, as these signals can be easily detected using wearable sensors \cite{s21041030}, making them convenient and non-invasive. Recently, there has been increasing interest in applying machine learning algorithms to classify pain intensity based on these signals allowing for a more objective and automated approach to pain assessment \cite{ozek2022review, s20020365, matsangidou2021machine, Werner2022}. They have also shown promising results in previous studies, which we discuss in \cref{sec:rel}. 
% {Section~\ref{sec:rel}} 

In this paper, we evaluate a novel transformer-encoder deep learning approach on EDA signals for automated pain intensity classification. The data from publically available BioVid dataset is utilized for the experiments. We aim to provide an objective, automatic, and convenient method of pain assessment that can be used in clinical settings and home settings.

%-------------------------------------------------------------------------
\begin{figure*}
	\centering
	\includegraphics[width=\linewidth]{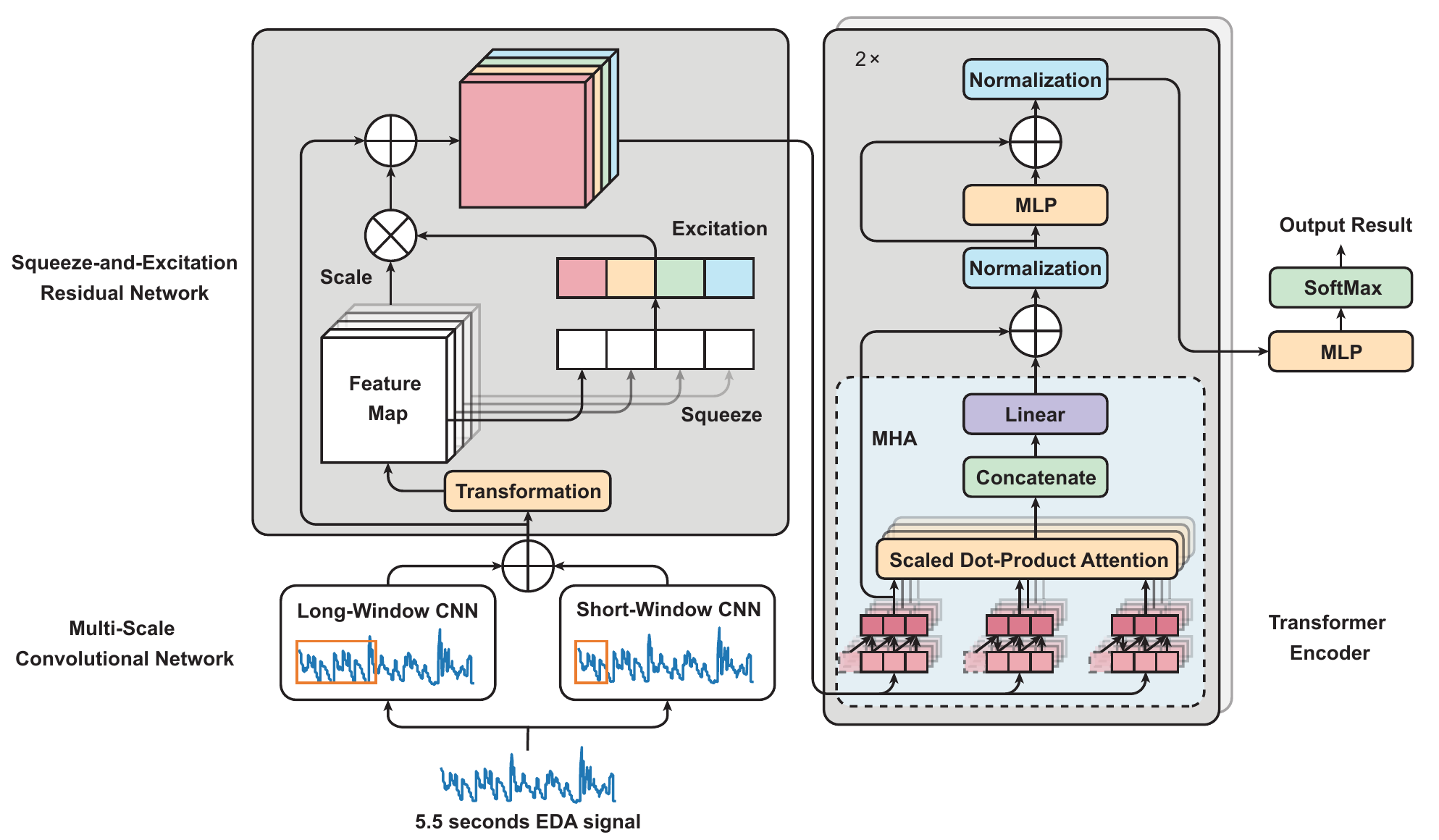}
% 	\vspace{-13pt}
    \caption{Outline framework of our proposed PainAttnNet. Left bottom: Multiscale Convolutional Network (MSCN). Left top: Squeeze-and-Excitation Residual Network (SEResNet). Right: Transformer Encoder.}
    \vspace{-10pt}
    \label{fig:outline}
\end{figure*}
%-------------------------------------------------------------------------

%-------------------------------------------------------------------------

\section{Related Work}
\label{sec:rel}

There has been growing interest in using conventional machine learning techniques, \eg support vector machine (SVM), k-nearest neighbors (KNN), regression model, bayesian model, and tree-based model, for the classification of pain intensity in order to improve the accuracy and efficiency of pain assessment based on physiological signals. 

Using an SVM and KNN to classify pain levels is a common strategy \cite{pouromran2021exploration, naeini2021pain, cao2021objective}. One study proposed an SVM model and used two separate feature selection strategies (univariate feature selection and sequential forward selection) during the feature extraction phase \cite{campbell2019feature}. In a similar fashion, one study identified low back pain based on features extracted from motion sensor data using an SVM model \cite{abdollahi2020using}.

A Bayesian network was utilized in one research to construct a decision support system to aid in the treatment of low back pain \cite{santra2020medical}. The system was capable of providing tailored therapeutic recommendations based on the unique characteristics of patient and medical history. One more common strategy is the use of random forests, which has been implemented in a variety of research projects. These studies \cite{werner2014automatic, kachele2015multimodal} applied random forest models to the BioVid Heat Pain Database \cite{biovid}, which included multidimensional datasets consisting of both video and physiological signals (\eg ECG, SCL, EMG, EEG). The other tree-based models, \eg AdaBoost, XGBoost, and TabNet, have also been applied in the classification of pain intensity. For example, in the study of Shi~\etal\cite{shi2022tree}, the researchers manually extracted features to categorize pain intensity using AdaBoost, XGBoost, and TabNet models. Similarly, Pouromran~\etal\cite{pouromran2021exploration} explored XGBoost for estimating pain intensity using catch22 \cite{lubba2019catch22} features of signals. Other studies implemented ADABoost and XGBoost \cite{naeini2021pain, cao2021objective} with filter-based feature selection methods, \eg gini impurity gain. 

Several studies have also integraded tree-based models with other machine learning techniques. For instance, Pouromran~\etal\cite{pouromran2022personalized} employed BiLSTM to extract the features which were then output to the XGBoost, resulting in high performance across four categories of pain intensity. The BiLSTM layer, which is an enhanced RNN with gates to govern the information flow, has the ability to tackle the problem of vanishing and exploding gradients in RNN. Wang~\etal\cite{wang2020hybrid} introduced a hybrid deep learning model with a BiLSTM layer to extract temporal features. They fused these with hand-crafted features, \eg mean, maximum, and standard deviation of SCL and fed to a multi-layer perceptron (MLP) block to classify the signals. Lopez-Martinez and Picard~\etal\cite{lopez2017multi} proposed a multi-task deep MLP to classify pain intensity using physiological signals, \eg heart rate variability, skin conductance to classify pain intensity. Similarly Gouverneur~\etal\cite{gouverneur2021comparison} applied MLP with unique hand-crafted features to classify heat-induced pain classification.

Thiam~\etal\cite{thiam2019exploring} proposed a deep learning model that utilizes a deep CNN framework followed by a block of fully connected layers (FCL) for the pain recognition. Similarly, Subramaniam and Dass~\etal\cite{subramaniam2020automated} built a hybrid deep learning model that combines CNN with LSTM for pain recognition. These authors used such a framework to extract temporal features from hand-picked samples of BioVid, and then used a FCL to classify the signals into pain or no pain categories. 

These models described above have demonstrated potential in pain intensity classifications, but they also have limitations. RNNs, despite their ability to capture temporal dependencies in sequential data, may struggle to maintain long-term dependencies in the input sequences. In addition, RNNs are not amenable for training in parallel due to their recurrent nature. On the other hand, MLPs tend to have a limited capacity to capture temporal dependencies of the input signals. CNNs have shown promising results in pain intensity classification, but may not be effective for modeling temporal dependencies among EDA data. To overcome these limitations, we propose PainAttnNet (PAN), a novel transformer-encoder deep-learning framework for classifying pain intensities using physiological signals as inputs.

In \cref{sec:imp}, we will provide a detailed implementation of our proposed model (See \cref{fig:outline}) to resolve the above issues. In \cref{sec:exp} we will introduce the dataset we used, experimental results, evaluation metrics, baseline models comparison and model analysis. We provide the discussion in \cref{sec:dis}

%------------------------------------------------------------------------

\begin{figure}
	\centering
	\includegraphics[width=\linewidth]{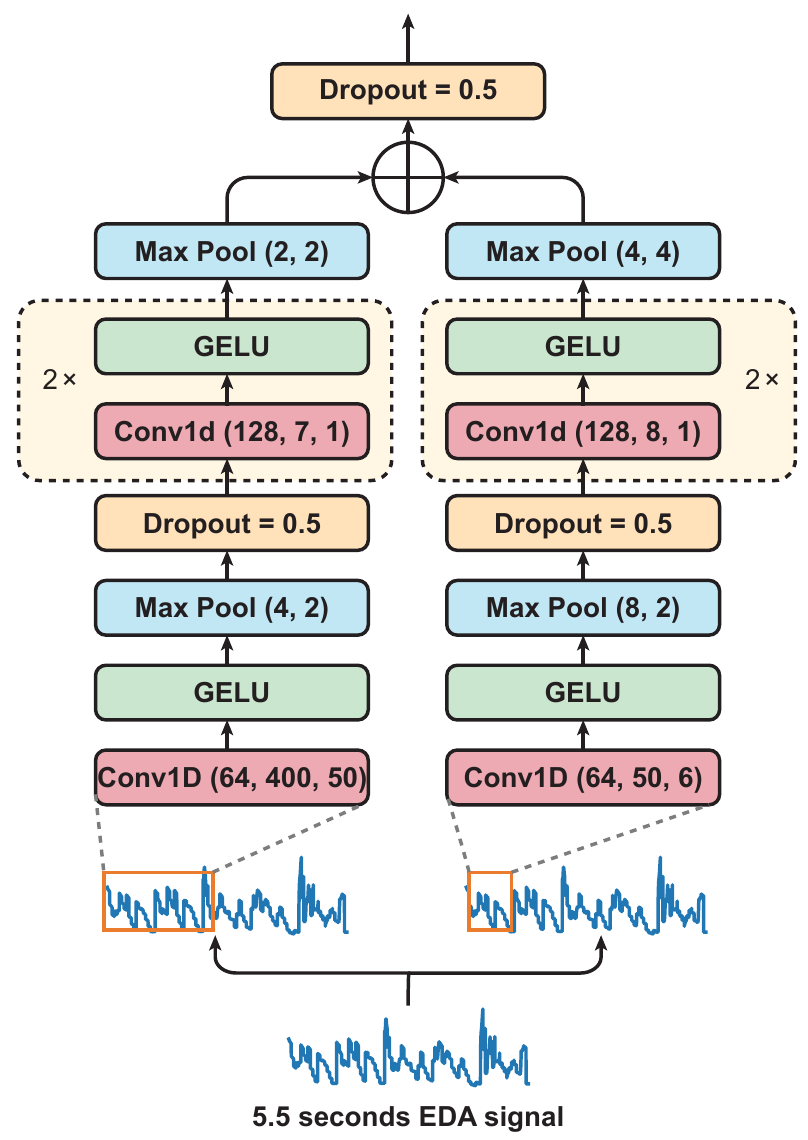}
% 	\vspace{-13pt}
    \caption{The structure of multiscale convolutional network (MSCN).}
    \vspace{-10pt}
    \label{fig:mscn}
\end{figure}

%------------------------------------------------------------------------

\section{Methodology}

Our implementation section includes five parts: (1) an overview of our framework, (2) Multiscale Convolutional Network for feature extraction, (3) Adaptive Recalibration Block for highlighting relevant features, (4) Temporal Convolutional Network for capturing temporal dependencies, and (5) Multi-head Attention mechanism for further improving performance.

\label{sec:imp}
\subsection{Outline of PainAttnNet}

To address the limitations of existing models for classifying pain intensity using physiological signals, we propose a novel framework that combines multiple networks and models (\cref{fig:outline}). Our framework aims to effectively classify pain intensity from physiological signals by utilizing various strategies to extract the features from the signals.

We initially employ a multiscale convolutional network (MSCN) to extract long- and short-window features from signals, \eg Electrodermal Activity (EDA). These extracted features can capture important information about the overall trend and variations in the signal, providing valuable insight into the pain intensity.

Next, we use Squeeze-and-Excitation Residual Network (SEResNet) to learn the interdependencies among the extracted features to enhance the representation capability of the features. SEResNet consist of two main components: a squeeze operation, which reduces the number of channels in the feature maps by taking their spatial average, and an excitation operation, which scales the channel-wise feature maps using a weighted sum of the squeezed features. This allows the network to selectively weight the importance of different channels and adaptively recalibrate the feature maps.

Finally, to capture the temporal representations of the extracted features, we use a multi-head attention mechanism in conjunction with a temporal (causal) convolutional network. The multi-head attention mechanism allows the network to attend to different parts of the input sequence simultaneously, and the temporal convolution network effectively captures the dependencies between the input and output over time. The mechanism behind the multi-head attention rests on the idea of scaling the dot product of the query and key vectors by the square root of their dimensionality, followed by a weighted sum of the values using the scaled dot products as weights. This mechanism allows the network to attend to different parts of the input sequence in a parallel fashion. On the other hand, the temporal convolution network uses an auto-regressive operation to effectively capture the dependencies between the sequence over time, while also allowing the end-to-end network training.

Overall, the proposed model aims to extract and analyze features from physiological signals comprehensively and effectively, improving the accuracy of pain intensity classification. In the next section, we will provide a detailed implementation of the proposed model.

\subsection{Multiscale Convolutional Network (MSCN)}

As EDA signals are inherently non-stationary. In the proposed approach, we employ a MSCN to effectively capture the various kinds of features from EDA signals (\cref{fig:mscn}). To accomplish this, the MSCN architecture is intended to sample varied lengths of EDA timestamps by utilizing two branches of convolutional layers, each with a different kernel size at the first layer. The first branch uses a kernel of 400 to cover a window of ~0.8 seconds while the second branch uses a kernel of 50 to cover a window of ~0.1 seconds, giving us a large segment and a small segment of features, respectively. The deep learning models presented in several studies \cite{peng2020multi, eldele2021sleep, li2016visual, gong2019cnn} inspired this technique. \cref{fig:mscn} depicts the network architecture, which consists of two max-pooling layers and three convolutions per branch, and the output of each convolutional layer is normalized by one batch normalization layer before being activated using Gaussian Error Linear Unit (GELU). Max-pooling, in particular, is a technique for downsampling an input representation, which reduces the dimensionality of the feature maps and controls overfitting. It is used to determine the maximum value of a certain feature map region. Given an input $\mathbf{X}=\{x_1,\ldots, x_N\}\in \bbR^{N\times L \times C}$, The operation of max-pooling can be described as:

\begin{equation}
  f_c(\mathbf{x})=\max_{i,j}(x_{i, j, c}).
  \label{eq:max}
\end{equation}
where $f$ is the output feature map, $\mathbf{x}$ is the input feature map per channel, $i$ and $j$ are the spatial dimensions and $c$ is the channel. The max pooling operation is applied to each channel separately, and the function $f_c(\mathbf{x})$ gives the maximum value of the elements in channel c. For example, $f_c(\mathbf{x})$ would be the maximum value of all elements in the $c-th$ channel of the feature map $\mathbf{X}$.

After each convolutional layer, the batch normalization layer accelerates network convergence by decreasing internal covariate shifts and stabilizes the training process \cite{ioffe2015batch}. Batch normalization normalizes the activations of the prior layer by using the channel-wise mean $mu_c$ and standard deviation $\sigma_c$. The batch normalization formulas are as follows: Let feature map $\mathbf{X} \in \bbR^{N\times L \times C}$ over a batch, where $L$ is the length of each feature, $N$ is the total number of features, and $C$ is the channel. The formula for batch normalization are as follows:

\begin{equation}
  y_{\gamma, \beta, c} = \frac{x_{i,j,c}-\mu_c}{\sigma_c}\cdot \gamma + \beta,
  \label{eq:batchNorm}
\end{equation}

here, 

\begin{equation}
  \mu_c = \frac{1}{NL}\sum_{i,j}x_{i,j,c},
\end{equation}

\begin{equation}
  \sigma_{c}^2 = \frac{1}{NL}\sum_{i,j}(x_{i,j,c}- \mu_c)^2. 
\end{equation}
where i and j d spatial indices and c is the channel index; $\mu_c$ and $\sigma_c^2$ are the mean of the values and the variance in channel $c$ for the current batch, respectively. In the above euqations, $\gamma$ and $\beta$ are learnable parameters introduced to allow the network to learn an appropriate normalization even when the input is not normally distributed.

GELU is a form of activation function that is a smooth approximation of the behavior of the rectified linear unit (ReLU)\cite{nair2010rectified} to prevent neurons from vanishing while limiting how deep into the negative regime activations \cite{Hendrycks2016gelu}. This allows having some negative weights to pass through the network, which is important to send the information to the subsequent task in SEResNet. As GELU follows the Batch Normalization Layer, the feature map inputs $ \mathbf{X} \sim \mathcal{N}(0,1)$. The GELU is defined as: 

\begin{equation}
  g(\mathbf{x}) := \mathbf{x} \cdot \Phi(\mathbf{x}) =\mathbf{x}\cdot \frac{1}{2}(1 + \mathbf{erf}(\frac{\mathbf{x}}{\sqrt{2}})). 
\end{equation}
where $\Phi(\mathbf{x})$ is the cumulative distribution function of the standard normal distribution ${P}\left(\mathbf{X}\leq\mathbf{x}\right)$. The $\mathbf{erf}(\cdot)$ is the error function. The ability of GELU is to boost the representation capabilities of the network by introducing a stochastic component that enables more diverse and ; GELU is one of the network's primary strengths. In addition, it has been demonstrated that GELU has a more stable gradient and a more robust optimization landscape than ReLU and leaky ReLU, because of this GELU can promote faster convergence and improved generalization performance.

Additionally, we employ a dropout layer after the first max pooling in both branches, and concatenate the output features from the two branches of the MSCN.  

\subsection{Squeeze-and-Excitation Residual Network (SEResNet)}

%------------------------------------------------------------------------

\begin{figure}
	\centering
	\includegraphics[width=\linewidth]{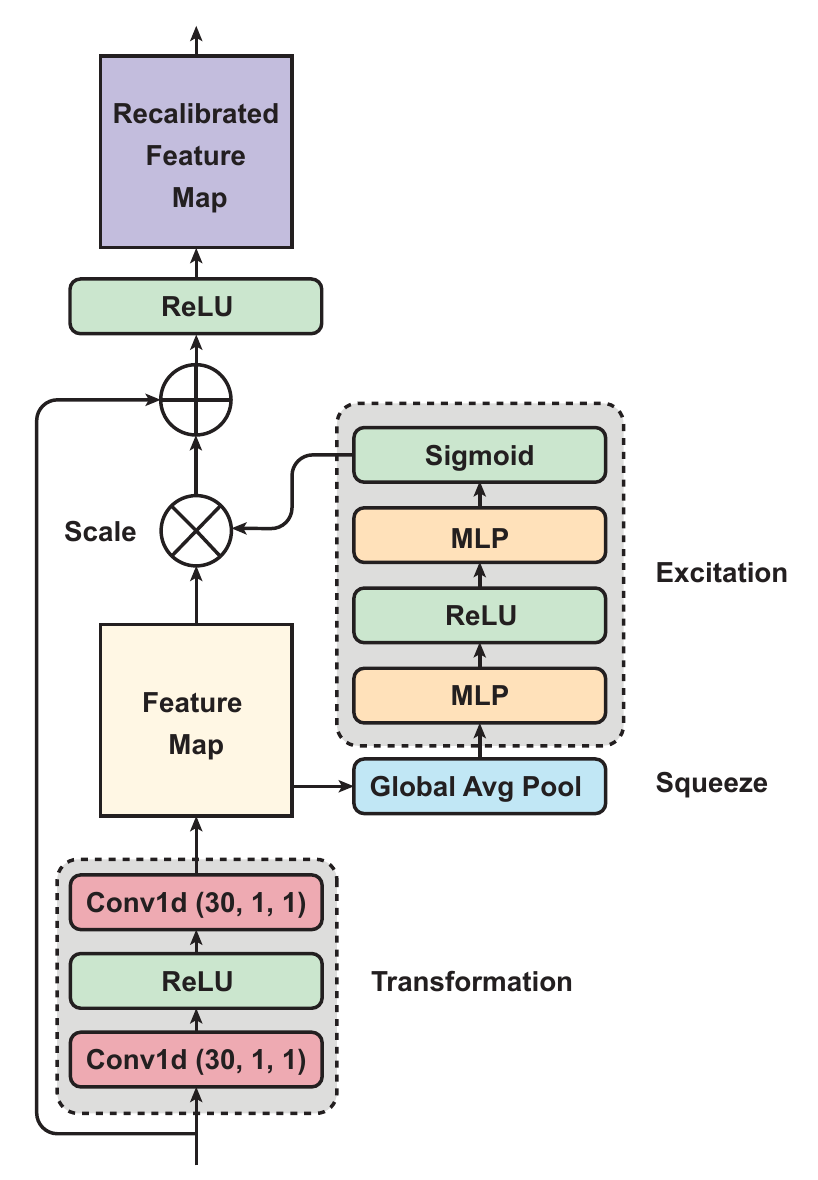}
% 	\vspace{-13pt}
    \caption{The outline of Squeeze-and-Excitation Residual Network (SEResNet).}
    \vspace{-10pt}
    \label{fig:senet}
\end{figure}

%------------------------------------------------------------------------

Using the SEResNet (\cref{fig:senet}), we can adaptively recalibrate the concatenated features from the MSCN to enhance the most important global spatial information of EDA signals. The mechanism of the SEResNet aims to model the inter-dependencies between the channels to enhance the convolutional features and increase the sensitivity of the network to the most informative features \cite{hu2018senet}, which is useful for our subsequent tasks. The SEResNet operates by compressing the spatial information of the feature maps into a global information embedding, and the excitation operation uses this descriptor to adaptively scale the feature maps (\cref{fig:senet}). Particularly, we initially employ two convolutional layers with a kernel and stride size of 1, and an activation of ReLU. Here we use ReLU, other than GELU, to improve the performance on the convergence. At the squeezing stage in the SEResNet, the global spatial information from the two convolutional layers are then compressed by global average pooling. It reduces the spatial dimension of feature maps while keeping the most informative features. Let the feature map from the MSCN as $\mathbf{X} \in \bbR^{N \times L \times C}$, we apply two convolutional layers to $\mathbf{X}$ and have new feature maps $\mathbf{V} \in \bbR^{N \times L \times C}$ shrink the $\mathbf{X}$ to generate the statistics $\mathbf{z}\in \bbR^{C}$: 

\begin{equation} 
  \textbf{z}_{c}=\frac{1}{NL}\sum\limits_{i=1}^{N}\sum\limits_{j=1}^{L}v_{i,j,c }, 
\end{equation}
where $\textbf{z}_c$ is the global average of $L$ data points per each channel. Next comes the excitation (adaptive recalibration) stage, in which two FCL generate the statistics used to scale the feature maps. As a bottleneck, the first FCL with ReLU is used to reduce the dimensionality of the feature maps. The second with sigmoid recovers the channel dimensions to their original size by performing a dimensionality-increasing operation. Let the $\mathbf{z} \in \bbR^C$. We define adaptive recalibration as follows:

\begin{equation} 
  \mathbf{\alpha}=\sigma(\mathbf{W}_{2}\delta(\mathbf{W}_{1}\mathbf{z})),
\end{equation}
where $r$ is the reduction ratio. $\delta$ denotes the ReLU function, and $\sigma$ refers to the sigmoid function. $\mathbf{W}_{1}\in \bbR^{\frac{C}{r}\times C} \ \text{and} \ \mathbf{W}_{2} \in  \bbR^{C\times \frac{C}{r}}$ is the learnable weights for the first FC layer and the second, respectively. These weights reveal the channel dependencies and provide information about the most informative channel.

Then the original feature map $\mathbf{v}$ is scaled by the activation $\mathbf{\alpha}$, and this is done by channel-wise multiplication: 

\begin{equation} 
  \mathbf{M}=\mathbf{\alpha}_{c} \otimes \mathbf{v}_{c}, 
\end{equation}

\begin{equation} 
  \tilde{\mathbf{X}}=\mathbf{X} \oplus \mathbf{M}.
\end{equation}
where $\tilde{\mathbf{X}}$ is the final output of the SEResNet, which results from the original input $\mathbf{X}$ and the enhanced features $\mathbf{M}$.

\subsection{Transformer Encoder}
\subsubsection{Temporal Convolutional Network (TCN)}

TCN framework, inspired by the studies of Lea~\etal\cite{lea2016temporal} and Van den Oord~\etal\cite{van2016conditional, oord2016wavenet}, has been used effectively for processing and generating sequential data, \eg audio or images. TCN employs one dimension convolutional layers to capture the temporal dependencies among the input data in a sequence, \eg previous recalibrated SEResNet features. In contrast to a regular convolutional network, the TCN's output at time $t$ depends only on the inputs before $t$. TCN only permits the convolutional layer to look back in time by masking future inputs. Like the regular convolutional network, each convolutional layer contains a kernel with a specific width to extract certain patterns or dependencies in the input data across time before the present $t$. To make input and output length the same, additional padding is added to the left side of input to compensate for the input's window shift. 

Let input feature map $\mathbf{X}\in \bbR^{1 \times L \times C_1}$, where $L$ is the input length, and $C_1$ is the dimension of input channels. We have kernel $\mathbf{W}\in \bbR^{K \times C_1 \times C_2}$, and the size of padding $(K-1)\in \bbR$, where $K$ is the size of the kernel, and $C_2$ is the dimension of output channels. Then we have the output from TCN as $\varphi(\cdot) \in \bbR^{1 \times L \times C_2}$. This approach can assist us in constructing an effective auto-regressive model that only retrieves temporal information with a particular time frame from the past without cheating by utilizing knowledge about the future.

% \begin{equation}
%   \varphi(\mathbf{x}) = \sum_{i=1}^{K}\sum_{j=1}^{C_1} x_{l+i-1} w_{i,j}
%   \end{equation}

\subsubsection{Multi-Head Attention (MHA)}

%------------------------------------------------------------------------

\begin{figure}
	\centering
	\includegraphics[width=\linewidth]{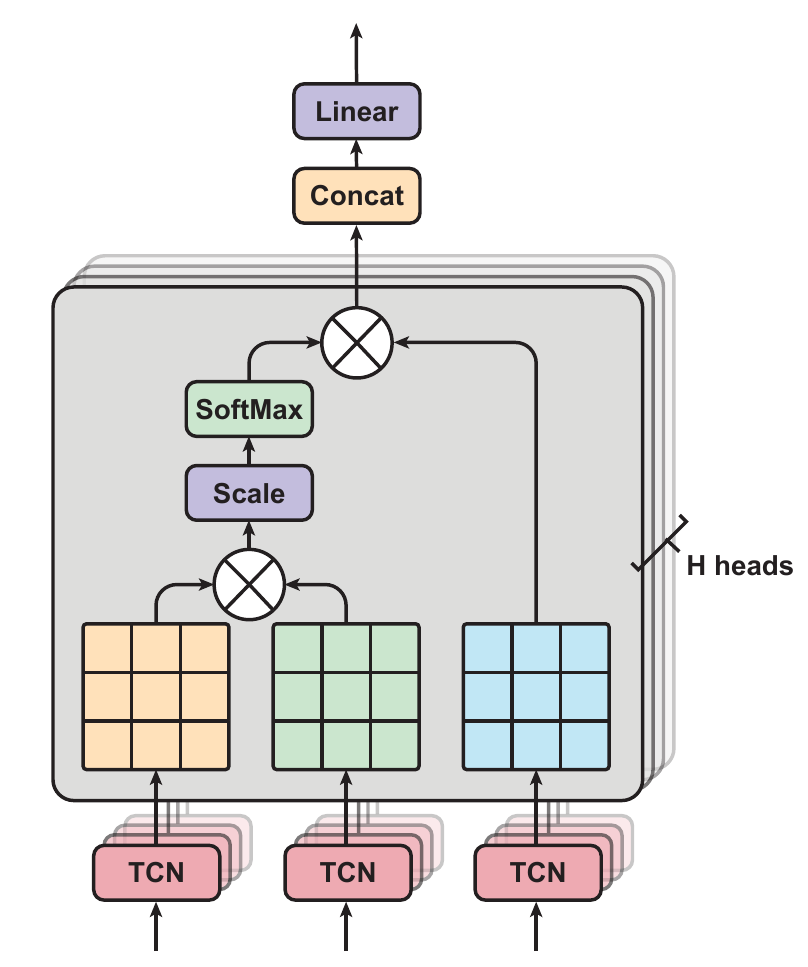}
% 	\vspace{-13pt}
    \caption{The structure of multi-head attention, consisits H heads of Scaled Dot-Product Attention layers with three inputs from TCNs.}
    \vspace{-10pt}
    \label{fig:mha}
\end{figure}
%------------------------------------------------------------------------

Multi-Head Attention (MHA) is the main part of the Transformer Encoder. It is a popular method for learning long-term relationships in sequences of features (\cref{fig:mha}). We adapt this algorithm from Dosovitskiy~\etal\cite{dosovitskiy2020image}, Vaswani~\etal\cite{vaswani2018attn}, and Bahdanau~\etal\cite{bahdanau2014neural}. It has significant performance in different fields, \eg GPT \cite{brown2020gpt} and BERT \cite{devlin2019bert} models in natural language process, and physiological signals classification for sleep Eldele~\etal\cite{eldele2021sleep}, Zhu~\etal\cite{zhu2020convolution}. MHA consists of multiple layers of Scaled Dot-Product Attention, where each layer is capable of learning different temporal dependencies from the input feature maps (\cref{fig:mha}). MHA aims to obtain a more comprehensive understanding of how the $i$th feature is relevant with $j$th features by processing them through multiple attention mechanisms. In particular, let the output feature maps from SEResNet, $\mathbf{X}=\{x_1, \ldots, x_N \} \in \bbR^{N \times L}$. Then we take three duplicates of $\mathbf{X}$ such that $\tilde{\mathbf{X}}=\varphi(\mathbf{X})$, where $\varphi(\cdot)$ is the function of TCN, and $\tilde{\mathbf{X}}$ is the output of TCN. Next we send the three outputs, $\tilde{\mathbf{X}}^{(Q)}, \tilde{\mathbf{X}}^{(K)}, \tilde{\mathbf{X}}^{(V)}$ to attention layers to calculate the weighted sum of the input, the attention scores $\mathbf{z}_i$: 

\begin{equation}
  \mathbf{z}_i = \sum_{j=1}^{L}\alpha_{ij}\varphi\left(\tilde{\mathbf{x}}_{j}^{(V)}\right),
\end{equation}
the weight $\alpha_{ij}$ of each $\varphi(x_j)$ is computed by:

% \begin{equation}
%   \alpha_{ij} = \text{softmax}\left(\frac{\tilde{\mathbf{X}}^{(Q)} \tilde{\mathbf{X}}^{(K)\top}}{\sqrt{L}}\right)
% \end{equation}

\begin{equation}
  \alpha_{ij}=\frac{\text{exp}(e_{ij})}{\sum_{r=1}^{L}\text{exp}(e_{ir})}, 
\end{equation}
here, 
\begin{equation}
  e_{ij} = \frac{1}{\sqrt{L}} \cdot \tilde{\mathbf{x}}_{i}^{(Q)} \cdot \tilde{\mathbf{x}}_{j}^{(K)\top}.
\end{equation}
then the output of one attention layer is $\mathbf{z} = \{z_0, \ldots, z_L  \} \in \bbR^{N \times L}$. 

Next, MHA calculates all the attention scores $\mathbf{Z}^{(H)}$ from multiple attention layers parallelly, and then concatenate them into $\tilde{\mathbf{Z}}_{\text{MHA}} \in \bbR^{N \times HL}$, where $H$ is the number of attention heads, and $HL$ is the overall length of the concatenated attention scores. 

We apply a linear transformation with learnable weight $W \in \bbR^{HL \times L}$ to make the input and output dimensions the same so that we can easily process the subsequent stages. The overall equation for MHA is as follows:

\begin{equation}
  \tilde{\mathbf{Z}}_{\text{MHA}}= \text{Concat}(\mathbf{z}^{(1)}, \ldots, \mathbf{z}^{(H)}) \cdot W \in \bbR^{N \times L}.
\end{equation}

After concatenating these attention scores, we process them with the original $\tilde{\mathbf{X}}$ using an addition operation and layer normalization adopted from \cite{ba2016layer}, formed as $\Phi(\tilde{\mathbf{X}} + \tilde{\mathbf{Z}}_{\text{MHA}})$, which can be described as a residual layer with layernorm funciton $\Phi_1(\cdot)$. The output of $\Phi_1(\cdot)$ is then passed through the following two fully connected networks and the second residual layer $\Phi_2(\cdot)$. Finally, the pain intensity categorization results are obtained from another two fully connected networks, which are then followed by a Softmax function.

%------------------------------------------------------------------------

\section{Experimental Results}
\label{sec:exp}

%------------------------------------------------------------------------

\begin{table}[!htp]
  \centering
  \begin{tabular}{lccc}
    \toprule
    Tasks & ACC & $MF_1$  & $\kappa$  \\
    \midrule
    $T_{0} \ \text{vs.} \  T_{1} \ \text{vs.} \  T_{2} \ \text{vs.} \  T_{3} \ \text{vs.} \  T_{4}$ & 34.46 & 37.19& 0.18\\
    $T_{0} \ \text{vs.} \ (T_{1}, T_{2}, T_{3}, T_{4})$ & 80.87& 78.32& 0.09\\
    $T_{0} \ \text{vs.} \  T_{1}$ & 54.63& 56.33& 0.09\\
    $T_{0} \ \text{vs.} \  T_{2}$ & 68.82& 69.04& 0.37\\
    $T_{0} \ \text{vs.} \ T_{3}$ & 76.93& 76.94& 0.53\\
    $T_{0} \ \text{vs.} \ T_{4}$ & \textbf{85.34} & \textbf{85.27} & \textbf{0.70}\\
    \bottomrule
  \end{tabular}
  \caption{PainAttnNet's performance through three evaluation metrics through six tasks: (1) $T_{0} \ \text{vs.} \  T_{1} \ \text{vs.} \  T_{2} \ \text{vs.} \  T_{3} \ \text{vs.} \  T_{4}$, (2) $T_{0} \ \text{vs.} \ (T_{1}, T_{2}, T_{3}, T_{4})$, (3) $T_{0} \ \text{vs.} \  T_{1}$, (4) $T_{0} \ \text{vs.} \  T_{2}$, (5) $T_{0} \ \text{vs.} \ T_{3}$, and (6) $T_{0} \ \text{vs.} \ T_{4}$, on BioVid dataset.}
  \label{tab:pan}
\end{table}

%------------------------------------------------------------------------

\subsection{BioVid Heat Pain Datase}

%------------------------------------------------------------------------

\begin{table*}[!htp]
  \centering
  \begin{tabular}{lccccc}
    \toprule
    Method & $T_{0} \ \text{vs.} \  T_{1}$ & $T_{0} \ \text{vs.} \  T_{2}$ & $T_{0} \ \text{vs.} \ T_{3}$ & $T_{0} \ \text{vs.} \ T_{4}$ & Procedure \\
    \midrule
    $\text{CNN + LSTM}^{\ddagger}$ \cite{subramaniam2020automated} & 85.65 & 74.47& 80.80 & 80.17 & 5.5 s Segmentation, $n = 67 \times 20 \times 5$\\
    $\text{CNN}^{\ddagger}$ \cite{thiam2019exploring} & 61.67 & 66.93 & 76.38 & 84.57 & 4.5 s Segmentation, $n = 87 \times 20 \times 5$\\
    \midrule
    Random Forest \cite{werner2014automatic}& 55.40 & 60.20 & 65.90 & 73.80 & \multirow{7}{*}{5.5 s Segmentation; $n = 87 \times 20 \times 5$}\\
    MT-NN \cite{lopez2017multi} & 50.01 & 60.34 & 69.76 & 79.98 &\\
    SVM \cite{pouromran2021exploration} & -  & - & - & 83.30 &\\
    TabNet \cite{shi2022tree} & $\textbf{65.57}$ & 67.76 & 74.54 & 83.99 &\\
    MLP \cite{gouverneur2021comparison} & 59.08 & 65.09 & 75.14 & 84.22 &\\
    XGBoost \cite{shi2022tree} & 61.49 & 68.39 & 76.15 & 85.23 &\\
    PainAttnNet (Ours) & 54.63 & $\textbf{68.82}$ & $\textbf{76.93}$ & $\textbf{85.34}$ &\\
    \bottomrule
  \end{tabular}
  % \begin{tablenotes}
  %   \small
  % \end{tablenotes}
  \caption{The performance comparison between PainAttnNet and other SOTA approaches. $\ddagger$: as these two approches proposed two different procedures on the data input, we just list them here but are not able to compare with others. }
  \label{tab:sota}
\end{table*}

%------------------------------------------------------------------------

%-------------------------------------------------------------------------
\begin{figure*}
	\centering
	\includegraphics[width=\linewidth]{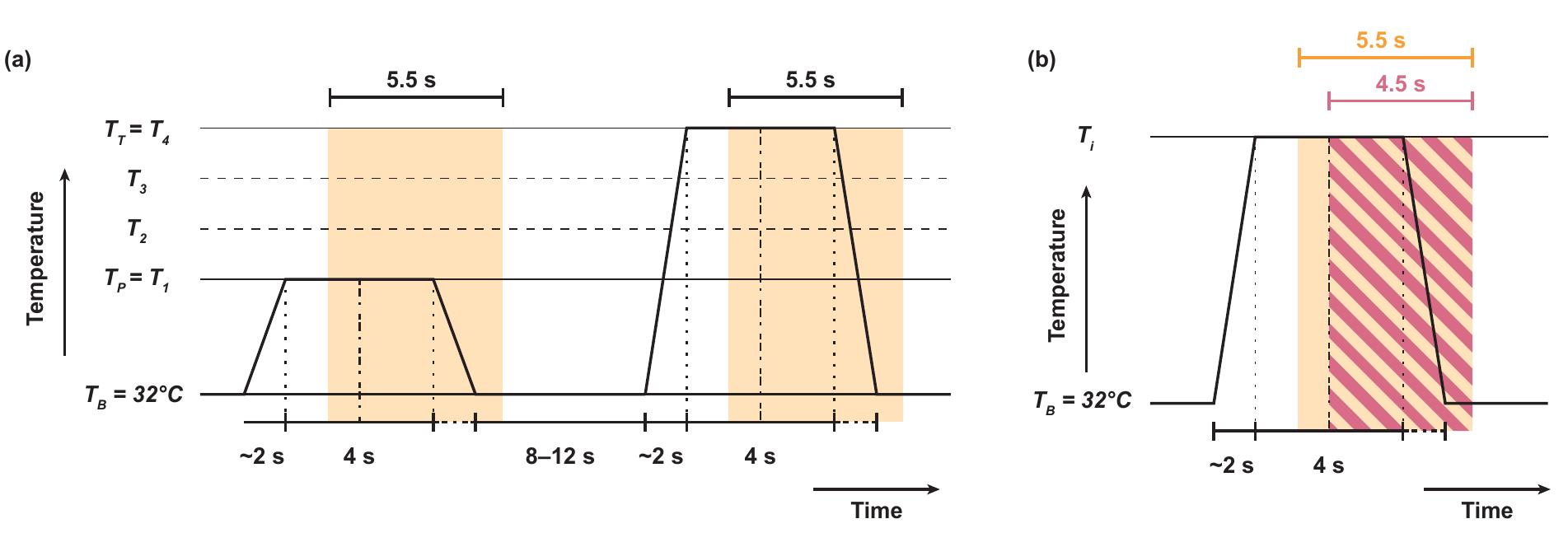}
% 	\vspace{-13pt}
  \caption{The heat stimuli, with a break in between interval and window segmentations. (a) demonstrates the original experiment settings of BioVid, with a duration of 4 seconds for each heat stimulus and an interval of 8 to 12 seconds between each stimulus. The yellow segmentation displays the 5-second timeframe for each collected signal. (b) Thiam~\etal introduces a different segmentation in red-strip rectangle which takes 4.5 seconds as opposed to 5.5 seconds.}
  \vspace{-10pt}
  \label{fig:biovid}
\end{figure*}
% \begin{figure*}[!thb]
%   \centering
%   \begin{subfigure}{0.55\linewidth}
%     % \fbox{\rule{0pt}{2in} \rule{.9\linewidth}{0pt}}
%     \includegraphics[width=\linewidth]{figures/plots/biovid.pdf}
%     \caption{An example of a subfigure.}
%     \label{fig:biovid}
%   \end{subfigure}
%   \hfill
%   \begin{subfigure}{0.38\linewidth}
%     % \fbox{\rule{0pt}{2in} \rule{.9\linewidth}{0pt}}
%     \includegraphics[width=\linewidth]{figures/plots/seg.pdf}
%     \caption{Another example of a subfigure.}
%     \label{fig:segmentation}
%   \end{subfigure}
%   \caption{Example of a short caption, which should be centered.}
%   \label{fig:biovid_data}
% \end{figure*}

%-------------------------------------------------------------------------

In our experiment, we used the Electrodermal Activity (EDA) signals from BioVid Heat Pain Database (BioVid), generated by Walter~\etal\cite{biovid}. As described in \cref{sec:intro}, Electrodermal Activity (EDA) is an useful indicator of pain intensity \cite{ledowski2009monitoring}. Walter~\etal\cite{biovid} conducted a series of pain stimulus experiments in order to acquire five distinct datasets, including video signals caputuring the subjects' facial expression, SCL (also known as EDA), ECG, and EMG. The experiment featured 90 participants in ages: 18-35, 36-50 and 51-65. Each group has 30 subjects, with an equal number of males and females. At the beginning of the experiment, the authors callibrated each participant's pain threshold by progressively raising the temperature from the baseline $T_0= 32^\circ  C$ to determine the temperature stages $T_P$ and $T_T$; here TP represents the temperature stages at which the individual began to experience the heat pain; TT is the temperature at which the individual experiences intolerable pain.  Then four temperature stages can be determined as follows:

\begin{equation}
  T_i =  \left\{
      \begin{array}{ll}
        T_P + [(i-1) \times \gamma] & i\in\{1, 2, 3, 4\}\\
        T_B & i = 0
      \end{array}
  \right.
\end{equation}
here, 
\begin{equation}
  \gamma = (T_T-T_P)/4 
\end{equation}
where $T_P$ and $T_T$ are respectively defined as $T_1$ and $T_4$. The individual received heat stimuli through a thermode (PATHWAY, Medoc, Israel) connected to the right arm for the duration of the experiment. In each trial, pain stimulation was administered to each participant for a duration of 25 minutes. In each experiment, they determined five temperatures, $T_{i\in\{0, 1, 2, 3, 4\}}$, to induce five pain intensity levels from lowest to highest. Each temperature stimulus was delivered 20 times for 4 seconds, with a random interval of 8 to 12 seconds between each application (\cref{fig:biovid}a). During this interval, the temperatures were kept at the pain-free ($32^\circ C$) level. EDA, ECG, and EMG were collected by the according sensors to a sampling rate of 512 Hz with a segmentation in a length of 5.5 seconds. Due to technical issues in the studies, three subjects were excluded, resulting in a final count of 87. Therefore, the training sample of each signal creates a channel with dimensions of 2816 $\times$ 20 $\times$ 5 $\times$ 87. Informed by the the previous \cite{werner2014automatic, lopez2017multi, pouromran2021exploration, shi2022tree, gouverneur2021comparison, shi2022tree}, we adopted the data from BioVid and used the EDA signal in a dimension of 2816 $\times$ 20 $\times$ 5 $\times$ 87 with a 5.5 second segmentation as the input in our experiment for pain intensity classification based on five pain labels. We also discovered that Subramaniam and Dass~\etal\cite{subramaniam2020automated} removed 20 out of 87 subjects, resulting 2816 $\times$ 20 $\times$ 5 $\times$ 67 training samples. In contrast Thiam~\etal\cite{thiam2019exploring} utilized a 4.5 seconds segmentation as opposed to the original 5.5 seconds (\cref{fig:biovid}b). In next sections, we will compare these latest state-of-the-art methods.

% \begin{figure*}
% 	\centering
%   \includegraphics[width=\linewidth]{figures/plots/biovid.pdf}
% 	% \vspace{-13pt}
%   \caption{Outline framework of our proposed PainAttnNet.}
%   \vspace{-10pt}
% \end{figure*}

% \begin{figure}
% 	\centering
%   \includegraphics[width=\linewidth]{figures/plots/seg.pdf}
% % 	\vspace{-13pt}
%   \caption{Outline framework of our proposed PainAttnNet.}
%   \vspace{-10pt}
% \end{figure}

\subsection{Evaluation metrics}

We utilized the accuracy (ACC), Cohen Kappa ($\kappa$) \cite{cohen1960coefficient}, and macro F1 score ($MF_1$) to analyze the performance of our PainAttnNet on the classification of pain intensity: 

\begin{equation}
  ACC = \frac{1}{Q}\sum_{i=1}^{K}TP_i.
\end{equation}

\begin{equation}
  MF_1 = \frac{1}{K}\sum_{i=1}^{K}F_1 ,
\end{equation}
here, 

\begin{equation}
  F_{1,i} = \frac{2 \times Precision_i \times Recall_i}{Precision_i + Recall_i},
\end{equation}

\begin{equation}
  Precision_i = \frac{TP_i}{TP_i + FP_i},
\end{equation}

\begin{equation}
  Recall_i = \frac{TP_i}{TP_i + FN_i}.
\end{equation}
% and
% \begin{equation}
%   \kappa = \frac{p_o-p_e}{1-p_e},
% \end{equation}
% here, 
% \begin{equation}
%   p_o = \frac{1}{Q}\sum_{i=1}^{K}TP_i, 
% \end{equation}
% \begin{equation}
%   p_e = \frac{1}{Q}\sum_{i=1}^{K}TP_i, 
% \end{equation}
where $TP_i$ ,$TN_i$, and $FN_i$ are the true positive, true negative, and false negative of each class. Here $K$ is the total number of classes, and $Q$ is the total number of samples in the training set.

\subsection{Experimental Settings}
In our study, we compared PainAttnNet with six baselines, Random Forest \cite{werner2014automatic}, MT-NN \cite{lopez2017multi}, SVM \cite{pouromran2021exploration}, TabNet \cite{shi2022tree}, MLP \cite{gouverneur2021comparison}, and XGBoost \cite{shi2022tree}. In contrast, we also listed other two models, CNN + LSTM \cite{subramaniam2020automated}, CNN \cite{thiam2019exploring}, with different segmentation and sample selections on the EDA signals as the input. 

We implemented 87-fold cross-validation for the BioVid dataset by splitting the subjects into 87 groups, therefore, each subject is in one group as a leave-one-out cross-validation (LOOCV). We trained on 86 subjects and tested on one subject with 100 epochs for each iteration. Ultimately, the macro performance matrices were computed by combining the projected pain intensity classes from all 87 iterations. We created PainAttnNet using Python 3.10 and PyTorch 1.13 on a GPU powered by an Nvidia Quadro RTX 4000. We configured the optimizer as Adam with the initial learning rate of 1e-03, a weight decay of 1e-03a, and batch size of 128 for the training dataset. PyTorch's default settings for Betas and Epsilon were (0.9, 0.999) and 1e-08. In the transformer encoder, we utilized five heads for multi-head attention structure, with each feature's size being 75.

% \begin{itemize}
%   \item Random Forest \cite{werner2014automatic}

%   \item MT-NN \cite{lopez2017multi} 
  
%   \item SVM \cite{pouromran2021exploration} 
  
%   \item TabNet \cite{shi2022tree} 
  
%   \item MLP \cite{gouverneur2021comparison} 
  
%   \item XGBoost \cite{shi2022tree}
  
%   \item CNN + LSTM \cite{subramaniam2020automated}
  
%   \item CNN \cite{thiam2019exploring}
% \end{itemize}

\subsection{Performance of PainAttnNet}

We conducted six experimental tasks: (1) $T_{0} \ \text{vs.} \  T_{1} \text \ {vs.} \  T_{2} \ \text{vs.} \  T_{3} \ \text{vs.} \  T_{4}$, (2) $T_{0} \ \text{vs.} \ (T_{1}, T_{2}, T_{3}, T_{4})$, (3) $T_{0} \ \text{vs.} \  T_{1}$, (4) $T_{0} \ \text{vs.} \  T_{2}$, (5) $T_{0} \ \text{vs.} \ T_{3}$, and (6) $T_{0} \ \text{vs.} \ T_{4}$, to evaluate PainAttnNet's performance on the BioVid dataset (\cref{tab:pan}). Among these tasks, we are most interested in tasks 2, 5, and 6, as in clinical trials it is essential to distinguish between pain and no pain (Task 2), and it is crucial for us to understand the distinctions between no pain and nearly intolerable pain (Tasks 5 and 6), to improve the quality of patient care.

The six classification tasks listed in the table evaluate the performance of the PainAttnNet on the BioVid dataset (\cref{tab:pan}). The first task, $T_{0} \ \text{vs.} \ T_{1} \ \text{vs.} \ T_{2} \ \text{vs.} \ T_{3} \ \text{vs.} \ T_{4}$, involves classifying pain intensity levels into five categories: no pain ($\ T_{0}$), low pain ($\ T_{1}$), medium pain ($\ T_{2}$), high pain ($\ T_{3}$), and nearly intolerable pain ($\ T_{4}$).

Task 2, $T_{0} \ \text{vs.} \ (T_{1}, T_{2}, T_{3}, T_{4})$, involves classifying pain intensity levels into two categories: no pain (T0) and any level of pain (T1, T2, T3, T4).

Tasks 3, 4, 5, and 6 classify pain intensity levels into two categories for each specific pain level. For example, in the third task ($T_{0} \ \text{vs.} \ T_{1}$), the classifier is trained to distinguish between no pain ($\ T_{0}$) and low pain ($\ T_{1}$). Similarly, in the fourth task ($T_{0} \ \text{vs.} \ T_{2}$), the classifier is trained to distinguish between no pain ($\ T_{0}$) and medium pain ($\ T_{2}$), and so on.

Among these, tasks 2, 5, and 6 are particular interesting as they involve classifying instances into two categories: no pain ($\ T_{0}$) and pain. Task 2 is important as it involves distinguishing between no pain and any level of pain, which is essential in clinical trials. Tasks 5 and 6 are important since they distinguish between no pain and nearly intolerable pain ($\ T_{3}$ and $\ T_{40}$), which is crucial for improving the quality of patient care.

The results show that the PainAttnNet model performed best on Task 6, with an of 85.34\% accuracy, a macro F1 score of 85.27\%, and a Cohen Kappa of 0.70. The model performed weakly on Task 2, with an accuracy of 80.87\%, a macro F1 score of 78.32\%, and a Cohen Kappa of 0.09. The performance on Tasks 3, 4, and 5 falls in between the performance levels for Task 2 and Task 6 varying levels of accuracy, macro F1 score, and Cohen Kappa.

We also compared PainAttnNet to other SOTA and the latest approaches on the pain intensity classification for the BioVid dataset (\cref{tab:sota}). To be easy to make a comparison we only select four out of the previous six classification tasks: $T_{0} \ \text{vs.} \  T_{1}$, $T_{0} \ \text{vs.} \  T_{2}$, $T_{0} \ \text{vs.} \ T_{3}$, and $T_{0} \ \text{vs.} \ T_{4}$.

The first two approaches, CNN + LSTM \cite{subramaniam2020automated} and CNN \cite{thiam2019exploring}, used different sample selections and data segmentation, respectively. Therefore, we just list the results in the table \cref{tab:sota} as references but without comparison to the rest.

The results in the table (\cref{tab:sota}) show that our proposed model outperforms other SOTA approaches. In particular, PainAttnNet achieved the highest accuracy for tasks $T_{0} \ \text{vs.} \ T_{3}$, and $T_{0} \ \text{vs.} \ T_{4}$, where the distinction between no pain and nearly intolerable pain is crucial. In task $T_{0} \ \text{vs.} \  T_{2}$, our model achieved a slightly higher accuracy compared to the best-performing SOTA approach (68.82 vs 68.39). In task $T_{0} \ \text{vs.} \  T_{1}$, Shi~\etal\cite{shi2022tree} have the highest accuracy. 

In conclusion, the results of this comparison demonstrate that our proposed model, PainAttnNet, is a promising approach for classifying pain levels in EDA signals.

% \subsection{Ablation Study}

% \subsection{Sensitivity Analysis}

\section{Conclusions}
\label{sec:dis}
PainAttnNet is a unique framework we developed to classify the severity of pain based on EDA signals (PAN). A multiscale convolutional network (MSCN) and an Sequeeze-and-Excitation Residual Network (SEResNet) based on the feature extraction from EDA signals performed by PainAttnNet. The multi-head attention architecture consists of a temporal convolutional network (TCN) for catching temporal dependencies and multiple Scaled Dot-Product Attention layers for understanding the relationship among input temporal features. The results of the experiments conducted on the BioVid database indicates that our model achieves better results compared to other state-of-the-art methods.

The results suggest that the PainAttnNet model performs well on tasks distinguishing between no pain from various pain levels, but there is room for improvement in its ability to differentiate different levels of pain intensities. Moving further, we aim to apply masked models and adaptive embedding to enhance the feature information from subspace on the labelled data. To be more realistic for potential future clinical practice, we will utilize contrastive learning with transfer learning on both huge unlabeled data and little chunks of labeled data to determine if it can still provide significant results.

%-------------------------------------------------------------------------

%%%%%%%%% REFERENCES
{\small
\bibliographystyle{ieee_fullname}
\bibliography{PainAttnNet}
}

\end{document}